\documentclass{article}

\PassOptionsToPackage{numbers, sort&compress}{natbib}
\usepackage{natbib}
\usepackage[dblblindworkshop, final]{neurips_2025}
\usepackage[utf8]{inputenc} 
\usepackage[T1]{fontenc}    
\usepackage{hyperref}       
\usepackage{url}            
\usepackage{booktabs}       
\usepackage{amsfonts}       
\usepackage{nicefrac}       
\usepackage{microtype}      
\usepackage{xcolor}         
\usepackage{graphicx}
\usepackage{amsmath}
\usepackage[noabbrev,capitalize]{cleveref}
\usepackage[compact]{titlesec}

\workshoptitle{Efficient Reasoning}

\title{Calibrated Reasoning: An Explanatory Verifier for Dynamic and Efficient Problem-Solving}

\author{
  Anisha Garg \\
  \text{anisha.garg@cerebras.net} \And
  Engin Tekin \\
  \text{engin.tekin@cerebras.net} \And
  Yash More \\
  \text{yash.more@cerebras.net} \And
  David Bick \\
  \text{david.bick@cerebras.net} \And
  Nishit Neema \\
  \text{nishit.neema@cerebras.net} \And
  Ganesh Venkatesh \\
  \text{ganesh.venkatesh@cerebras.net} \And
  \textsc{Applied AI Research, Cerebras}
}

\begin{document}

\newcommand{\ignore}[1]{}
\newcommand{\fixme}[1]{\textcolor{red}{#1}}
\newcommand{\commentgv}[1]{\textcolor{orange}{#1}}
\newcommand{\commentresolved}[1]{\textcolor{blue}{#1}}
\maketitle


\begin{abstract}

Advanced test-time computing strategies are essential for scaling reasoning models, but their effectiveness is capped by the models' poor self-evaluation. We propose a pairwise Explanatory Verifier, trained via reinforcement learning (GRPO), that produces calibrated confidence scores and associated natural language reasoning for generated solutions. Our verifier improves the accuracy and efficiency of test-time strategies like best-of-n and self-reflection. Crucially, it excels at identifying challenging failure modes, such as when both candidate solutions are identically incorrect, succeeding where standard methods like majority voting fail.

\ignore{We propose a reinforcement learning–trained pairwise verifier, optimized with GRPO on Qwen-8B, that produces calibrated confidence scores and accurately detects incorrect answers even if they are both identical. It boosts accuracy across best-of-n, self-reflection, and single-shot settings while lowering computational cost. Crucially, it succeeds where majority-voting fails, showing that lightweight verifier training after reasoning-focused RL unlocks new capabilities and strengthens both verification and generation.}
\end{abstract}

\vspace{-0.7em}

\section{Introduction}
\label{sec:intro}

\ignore{
\begin{itemize}
    \item Turbocharging Reasoning models to solve ever more complex problems needs capable test-time orchestration that decomposes task, explores multiple trajectories, and build on what works.
    \item Challenge here is that reasoning models tend to be biased towards using a small number of strategies and are not good at judging their own responses. Esp a documented behavior is of the model being biased towards its own responses and hence accepting wrong responses. This limitation impacts model's ability to address really complex challenges by not being able to select when it got it correct vs when it needs to keep trying alternatives.
    \item In this work, we address this challenge by training a verifier using Reinforcement Learning. In particular, keeping the test-time compute in mind, the verifier tasks as input multiple executions and grades each one as both incorrect, both correct or identifies the correct one. Having two responses, helps the model spot inconsistencies and do a better job of identifying mistakes.
    \item Results show that our verifier is significantly better than the baseline reasoning model at selecting correct responses. Furthermore, it produces meaningful rationale along with the final rating which can further boost AI system's accuracy using test-time compute.
\end{itemize}
}

Advanced test-time strategies like multi-path exploration are key to solving increasingly complex problems~\cite{openai_o3_2025,openai_deep_research_2025, fu2025deepthinkconfidence}. However, their effectiveness is capped by a core limitation in reasoning models. These models struggle with reliable self-evaluation~\cite{huang2024largelanguagemodelsselfcorrect} and are often biased towards a narrow set of approaches (\cref{fig:distinct_answers}). This critical failure to discern correctness creates a bottleneck, preventing dynamic exploration of alternatives and hindering progress on scaling AI systems to address challenging tasks.

To overcome this bottleneck, we introduce an Explanatory Verifier trained via Reinforcement Learning~\cite{shao2024deepseekmathpushinglimitsmathematical} to provide both a calibrated judgment and a natural language rationale. Rather than assessing solutions in isolation, our verifier takes motivation from prior work~\cite{deepseekrm} to perform a more efficient relational analysis on pairs of reasoning trajectories to identify subtle errors and judge their correctness. This justified comparative judgment framework is designed to directly enhance common test-time reasoning strategies like best-of-n sampling~\cite{goodfellow_DL_textbook} and self-reflection~\cite{madaan2023selfrefineiterativerefinementselffeedback}.

To our knowledge, this is the first work to systematically train and analyze a pairwise, explanatory verifier to scale a complex reasoning system. Downstream evaluations on challenging benchmarks demonstrate that the verifier significantly improves accuracy in best-of-N sampling and self-reflection settings, often while using fewer computational resources. The foundation of this success is the verifier's robust calibration; our analysis reveals that it can reliably detect incorrect answers even in ambiguous scenarios where both candidates are wrong or identically flawed. These are precisely the cases where voting strategies fail. This reliability enables a more dynamic inference paradigm, where computationally expensive exploration is reserved only for problems where verifier confidence is low. Overall, our approach enables efficient scaling of reasoning models, providing a blueprint for dynamic systems that can tackle increasingly complex tasks with proportional resource allocation.


\ignore{Reasoning-augmented large language models (LLMs) have dramatically improved the ability of AI systems to solve complex problems, yielding substantial gains in accuracy on challenging mathematics and coding benchmarks. Despite these advances, prior work has revealed notable limitations of reasoning-focused reinforcement learning (RL) training. Specifically, such models often exhibit biases toward particular problem-solving strategies and show weaknesses in following instructions reliably. As a result, test-time scaling techniques, such as best-of-n sampling and self-reflection, achieve only limited improvements in solving difficult tasks. A fundamental constraint of even the most capable reasoning models is their inability to reliably verify their own answers, frequently overestimating correctness. While significant effort has gone into training reward models to produce scalar feedback scores (skywork-reward, meta-J1, ), evaluated primnarily on reward benchmarks, pairwise verifiers are typically evaluated on datasets where at least one candidate answer is guaranteed to be correct. 

In this paper, we propose a method to train a verifier using reinforcement learning, enabling efficient test-time scaling even on state-of-the-art reasoning models. To the best of our knowledge, no prior work has explored the use of a verifier for test-time scaling in reasoning tasks. We address this gap by training a pairwise verifier using GRPO on Qwen-8B. Our verifier is calibrated, such that its output reflects confidence in its judgment, and can identify cases where both candidate answers are incorrect. 

Analysis of the verifier’s ratings shows that it can reliably detect incorrect answers, even when both candidates are wrong or identical, where majority-vote strategies fail. This enables dynamic systems to apply computationally expensive best-of-n strategies only when verifier confidence is low. Downstream evaluations on reasoning benchmarks demonstrate that the verifier improves accuracy in both best-of-n and self-reflection settings while using fewer resources. Furthermore, verifier feedback enhances generation, and even in single-shot settings, it outperforms the baseline, showing that verifier training can directly improve generation quality as well. 

Overall, our approach enables efficient scaling of reasoning models, providing a blueprint for dynamic systems that can tackle increasingly complex tasks with minimal resource overhead.}
\section{Explanatory Verifier Training}
\label{sec:verifier}

\ignore{
\begin{itemize}
    \item Problem formulation in equations
    \item Data curation for verifier training
    \item Stages of training -- context length, dataset sampling, ...
    \item Implementation details and hyperparams
    \item Analysis of behavior shifts from RL training (goes into results)
\end{itemize}
}

\paragraph{Problem Formulation} We frame training the explanatory verifier as a reinforcement learning problem. The goal is to learn a policy $\pi$ that, given a problem instance consisting of a question $Q$ and two candidate responses $(R_A, R_B)$ \cite{liu2025inference}, generates a completion containing reasoning within <think> tags and ratings $V = (v_A, v_B) \in [0,10]$ in final answer. Ground-truth labels $y = (y_A, y_B) \in {0,1}$ indicate correctness, but the continuous output scale allows the model to express uncertainty in its judgments. Completion quality is measured by a reward function, $R(c, x)$, which is defined as the binary cross-entropy \cite{goodfellow_DL_textbook} between the normalized ratings and labels (see Section~\ref{sec:rewardshaping}). The training objective, optimized using Group Relative Policy Optimization (GRPO)~\cite{shao2024deepseekmathpushinglimitsmathematical}, is to find the optimal policy $\pi^*$ that maximizes the expected reward over the data distribution $\mathcal{D}$:$$ \pi^* = \underset{\pi}{\arg\max} \; \mathbb{E}_{x \sim \mathcal{D}, c \sim \pi(\cdot|p(x))} [R(c, x)] $$

\subsection{Training Dataset}
\label{sec:data}

The foundation of our verifier is a curated dataset derived from Numina Math~\cite{numina_math_datasets}, CodeForces, and LeetCode \cite{li2023taco, xia2025leetcodedatasettemporaldatasetrobust}. For each problem, we generated multiple solution attempts with Qwen3-8B \cite{qwen3technicalreport} to get correct and incorrect reasoning paths for training.

The base datasets contain many corner cases that are challenging for automated verification approaches. We implemented a rigorous curation process, detailed in Appendix \ref{sec:datapreparation}, so that our data contains only problems where automated verification is reliable to provide high-quality signal. We removed any problems that: had open-ended responses such as proof-based questions; contained multiple sub-questions and answers; were ambiguous or under-specified as determined by a strong LLM-as-a-judge; or whose final answer evaluated to anything other than a single numeric expression.    


Finally, our pairwise input format imposes a substantial constraint on context length. We remove the content within <think> tags, and further filter out samples that exceed 6,144 input tokens. In the end, we have 3,634 unique input tuples, $(Q, R_A, R_B)$, spanning 628 distinct questions, $Q$, for math dataset. We hold out 294 tuples for validation set.



\ignore{\subsection{Stages of Verifier Training}
\label{sec:stages}

We begin training with a math-only dataset and a maximum sequence length (MSL) of 8,192, following the recipe in~\cite{Nemotron think}. To improve data efficiency, we enable dynamic filtering of samples during training, discarding rollouts with no variance. During the first 100 training steps, the average generation length decreases from 4,000 to 2,500 tokens, while validation accuracy rises from 0.32 to 0.49, indicating that the model is not only learning to reason more effectively but also doing so more efficiently.

After continuing training to a total of 240 steps, we subsample the math dataset and incorporate a coding dataset curated using a similar process as described in ~\ref{sec:data}. With 120 additional steps of continued training, validation accuracy on coding problems improves from 0.32 to 0.43, while accuracy on math tasks is maintained.

When introducing the coding dataset, however, we observed entropy collapse at first, verifier ratings concentrate around the midpoint (score $\approx$ 5) at an 8k MSL, suggesting the model is struggling with problem difficulty. To address this, we filter out problems with difficulty greater than 0.8 and increase the MSL to 16k, which yields a stable training run.}

\subsection{Reward Shaping}
\label{sec:rewardshaping}

We train the verifier to produce ratings on a continuous scale from 0 to 10, where 0 indicates high confidence that the response is incorrect and 10 indicates high confidence that it is correct. A key challenge in our training was incentivizing the verifier to use the full [0, 10] rating scale despite a binary ground-truth signal ($y \in {0, 1}$). Mean squared error often pushes predictions to the extremes, so we designed a reward based on a variant of binary cross-entropy to encourage calibrated outputs. First, we normalize the model's raw rating for a given response, $v \in [0, 10]$, to a probability-like value $p = v/10$ and then clamp it to $[0.1, 0.9]$ to ensure training stability and prevent the logarithm from producing excessively large or infinite values: $\hat{p} = 0.1 + 0.8 \cdot p = 0.1 + 0.08 \cdot v$.
This transformed value $\hat{p}$ is then used to calculate a reward based on binary cross-entropy. For a completion $c$ with responses $A$ and $B$, reward is calculated as the sum of the rewards for each individual judgment:
$$R_{\text{shaped}}(c, x) = \underbrace{\left[ y_A \log(\hat{p}_A) + (1-y_A) \log(1-\hat{p}_A) \right]}_{\text{Reward for response A}} + \underbrace{\left[ y_B \log(\hat{p}_B) + (1-y_B) \log(1-\hat{p}_B) \right]}_{\text{Reward for response B}}$$
In this formulation, while clamping ensures numerical stability, the logarithmic formulation penalizes confident errors sharply and allows nuanced predictions without excessive penalty.

\ignore{The motivation for this formulation is twofold. First, the clamping ensures stability by preventing the `log' inputs from approaching zero. Second, the logarithmic nature of the reward function provides a desirable non-linear penalty. The reward gradient is steep for predictions that are confidently wrong (e.g., a low rating for a correct answer), thereby severely penalizing egregious errors. Conversely, the gradient is much gentler for predictions that are close to the correct label, giving the model the freedom to express nuanced uncertainty across the full rating scale without a major penalty.}

\subsection{Implementation Details}
\label{sec:impl}

We use GRPO~\cite{shao2024deepseekmathpushinglimitsmathematical} to train \textsc{Qwen3-8B}, following the VeRL implementation\footnote{https://github.com/volcengine/verl}, in a multi-stage manner on math and coding datasets~\cite{chen2025acereason}. The learning rate was $1\times10^{-6}$ with a 10-step linear warmup, KL coefficient 0.001, and rollouts with temperature and top\_p set to 1.0. Training is strictly on-policy, with one gradient step per rollout group. In \textbf{Stage 1}, we train on a math-only dataset with a maximum sequence length (MSL) of 8,192, generating 16 rollouts per prompt with a global batch size of 256. We use dynamic filtering to discard rollouts with no variance. Stage 1 proceeds for 240 steps. In \textbf{Stage 2}, training is extended with a curated coding dataset combined with a subset of math. We observe truncated responses, entropy collapse, and verifier ratings concentrated around the midpoint (score $\approx 5$), suggesting the model was struggling with problem difficulty within the given token budget. To mitigate these issues, MSL is increased to 16,384, global batch size to 512, an entropy coefficient of 0.001 is introduced, and high-difficulty problems (difficulty $>0.8$) are filtered using pass@k. Stage 2 continues for 120 steps.

\ignore{During the first 100 training steps, the average generation length decreases from 4,000 to 2,500 tokens, while validation accuracy rises from 0.32 to 0.49, indicating that the model is not only learning to reason more effectively but also doing so more efficiently. Stage 1 training was carried out for 240 steps.}
\ignore{With 120 additional steps of continued training, validation accuracy on coding problems improves from 0.32 to 0.43, while accuracy on math tasks is maintained.}
\section{Evaluating the Explanatory Verifier: Ability to discern, Downstream Performance, and Emergent Capabilities}
\label{sec:analysis}

We comprehensively evaluate our Explanatory Verifier on (i) its judgment accuracy, (ii) its enhancement of downstream reasoning tasks, and (iii) its emergent generative skills.

\subsection{Benchmarking the Verifier's Ability to Discern}
\label{sec:discern}

\begin{figure}[htb]
\centering
\includegraphics[width=0.8\textwidth,keepaspectratio]{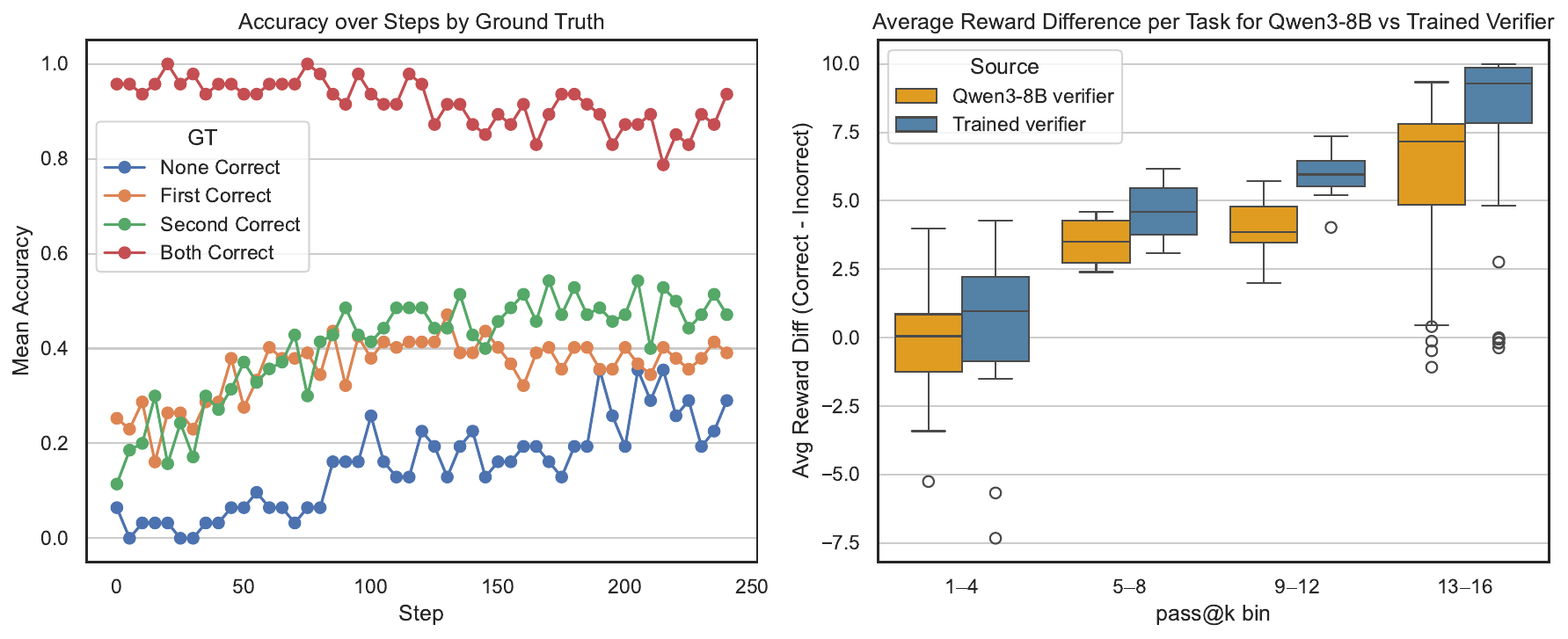}
\vspace{-8pt}
\caption{\textbf{(Left)} Training progression of the verifier's accuracy, broken down by the ground-truth (GT) configuration of the input pair. 
\textbf{(Right)} A comparison of the trained verifier (RL) against the baseline on generations from AIME 2024. The RL-trained verifier exhibits a much larger separation in ratings between correct and incorrect responses and shows lower variance in its predictions.}
\label{fig:verifier_analysis}
\vspace{-8pt}
\end{figure}

This section quantifies the judgment accuracy of Explanatory Verifier, benchmarking it against the self-evaluation capability of the baseline reasoning model. The results, summarized in Figure~\ref{fig:verifier_analysis}, demonstrate a significant improvement in the model's discerning abilities throughout its training.

\paragraph{Improvements in Discernment} Figure~\ref{fig:verifier_analysis} (Left) shows that, on a held-out validation set, the model progressively improves at evaluating correctness across all judgment scenarios, including identifying the better response and recognizing when both are incorrect. Initially (step 0), performance is highly skewed towards giving a high score to everything, but training develops reliable judgment in all settings. Of particular interest is the model's ability to identify when both inputs are incorrect.

\paragraph{Verifier Response Calibration}
Figure~\ref{fig:verifier_analysis} (Right) shows that the verifier's ratings become significantly more calibrated and consistent, beyond accuracy. This trend towards higher precision and lower variance holds true across problems of varying difficulty levels (grouped by pass@k bins).

Refer to  Appendix~\ref{sec:verifieranalysis} for more analysis on the underlying behavior of the verifier. Owing to these properties, we observe improvements when leveraging the verifier in test-time scaling, as discussed next. All results are reported using greedy decoding.

\subsection{Improving Token Efficiency in Best-of-N Sampling}
\label{sec:bestofn}

\begin{figure}[htb]
\centering
\includegraphics[width=0.9\textwidth,keepaspectratio]{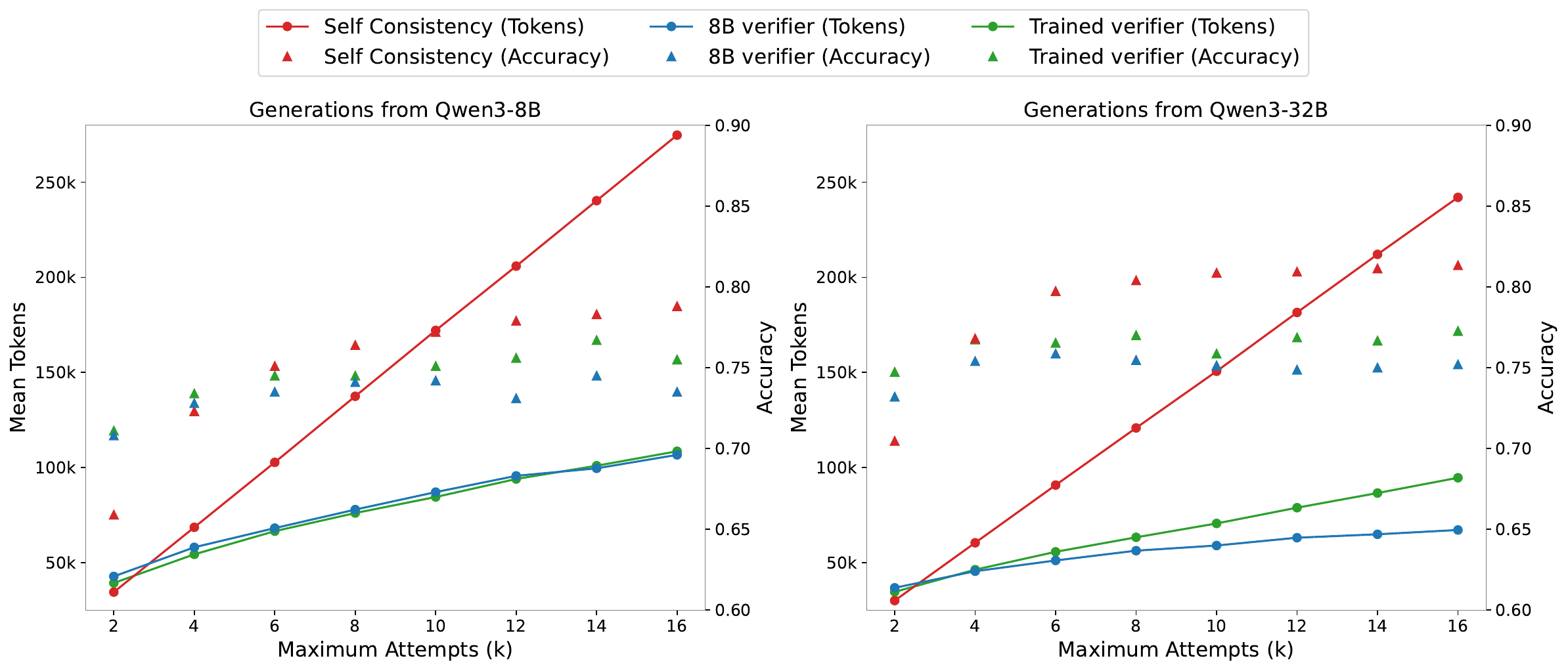}
\vspace{-7pt}
\caption{Across different values of k, the trained verifier consistently achieves higher accuracy than Qwen3-8B and provides a significant boost to the performance over self-consistency at lower k. Importantly, it requires substantially fewer tokens than self-consistency demonstrating both efficiency and stable performance across scales.}
\label{fig:verifier_retry} 
\vspace{-10pt}
\end{figure}

We evaluate the verifier as a retry mechanism during inference-time scaling. As shown in \cref{fig:verifier_retry}, the verifier achieves higher accuracy at lower values of $k$ compared to self-consistency, and comparable accuracy at higher $k$, while requiring 1--3$\times$ fewer tokens. Here, $k$ is the maximum number of candidate answers per task. In self-consistency, the final answer is chosen by majority vote over $k$ generations. In verifier-based approach, two candidates are scored first; additional generations are produced only if both are deemed incorrect, continuing until a correct answer is found or $k$ is reached. 

Although trained on \textsc{Qwen3-8B} outputs, the 8B verifier effectively evaluates larger models. For example, on AIME 2025, incorporating the verifier with Qwen3-32B generations achieves 0.77 accuracy while using only 75\% of the tokens compared to self-consistency.

\subsection{Improving Self-Reflection with Verifier-Guided Feedback}
\label{sec:cepolite}

Beyond judging correctness, the verifier’s natural language reasoning provides valuable feedback for iterative self-reflection (\cref{tab:verifier_cepo_results}). We generate two candidate answers with \textsc{Qwen3-8B}, score them with the verifier, and produce a final answer guided by its reasoning. This approach improves accuracy by up to 6 percentage points on \textsc{AIME} 2024 and 2025 compared to generating an answer using simple consistency check. Notably, gains are larger when using \textsc{GPT-OSS-20B} to generate final answer, likely due to its stronger instruction-following and reasoning. Beyond math, verifier feedback also boosts coding performance, demonstrating early applicability to this growing domain.

\begin{table}[h]
\centering
\caption{Accuracy of models with and without feedback from a verifier.}
\begin{tabular}{lcc|cc|cc}
\toprule
\textbf{Model} & \multicolumn{2}{c}{\textbf{AIME 2024}} & \multicolumn{2}{c}{\textbf{AIME 2025}} & \multicolumn{2}{c}{\textbf{LCB v6} \footnotesize{(01/01 - 08/01)}} \\
\cmidrule(lr){2-3} \cmidrule(lr){4-5} \cmidrule(lr){6-7}
                & W/o Verifier & W/ Verifier & W/o Verifier & W/ Verifier & W/o Verifier & W/ Verifier \\
\midrule
Qwen3-8B        & 0.77 & 0.77 & 0.67 & 0.68 & 0.39 & 0.44 \\
GPT-20B         & 0.77 & 0.80 & 0.65 & 0.71 & 0.49 & 0.49 \\
Qwen3-32B       & 0.77 & 0.76 & 0.67 & 0.69 & --   & --   \\
\bottomrule
\end{tabular}
\label{tab:verifier_cepo_results}
\vspace{-8pt}
\end{table}

\subsection{Emergent Generative Capabilities of the Verifier}
\label{sec:verifiergen}

A particularly promising finding is the emergent generative capability of the Explanatory Verifier. In a single-shot generation comparison against the baseline model, the verifier achieves statistically similar pass@1 accuracy (\cref{tab:aime_results}). This indicates that the intensive training for critical evaluation does not degrade, and possibly even enhances, the model's core reasoning abilities. This result paves the way for a new training paradigm that co-optimizes reasoning models for generating diverse solutions while accurately verifying their correctness, making them better suited for test-time scaling.

\begin{table}[h]
\caption{Performance on \textsc{AIME} 2024 and 2025 at $n\_repeat = 15$}
\centering
\begin{tabular}{lcc}
\toprule
\textbf{Benchmark} & \textbf{Qwen3-8B} & \textbf{Verifier} \\
\midrule
AIME 2024 & 0.77 & 0.78 \\
AIME 2025 & 0.66 & 0.68 \\
\bottomrule
\end{tabular}
\label{tab:aime_results}
\vspace{-8pt}
\end{table}

\section{Conclusion And Future Work}
\label{sec:conclusion}

\ignore{
\begin{itemize}
    \item Paper shows RL can be used to capable explanatory verifiers than can enhance test-time strategies ability to leverage reasoning models
    \items Looking forward -- exciting opportunities to enhance RL training loop with richer feedback than current sparse rewards as well as co-designing of reasoning models, verifiers and test-time strategies
    \item Next leap in AI system breakthrough will come up from effective agentic flows powered by multi-faceted models and our work is a step in that direction.
\end{itemize}
}

In this work, we demonstrated that test-time strategies can be significantly enhanced by using an Explanatory Verifier, trained with reinforcement learning, to overcome the core self-evaluation bottleneck of reasoning models. This approach opens promising avenues for future research, from training verifiers for test-time strategies using natural language feedback to the holistic co-design of integrated generator-verifier models. Ultimately, our work is a foundational step towards the next generation of AI systems: efficient, agentic systems where multi-faceted models can autonomously tackle problems of increasing complexity.

\ignore{We have shown that specialist verifier models can amplify the ability of reasoning-augmented LLMs to solve complex problems, enabling more efficient and accurate use of test-time compute. Our results highlight that reinforcement learning flows for reasoning hold promise not only for training stronger reasoning models, but also for developing specialist capabilities that dynamically guide inference. Looking forward, we see exciting opportunities to extend this line of work: incorporating consistency losses to ensure stable learning regardless of answer order, expanding training beyond mathematics to coding and puzzle-solving tasks, and refining reward functions for near-perfect calibration. More broadly, these directions suggest a pathway toward dynamic systems that combine reasoning, planning, and task decomposition, capable of solving increasingly complex problems through flexible exploration of the solution space.}
\bibliographystyle{unsrtnat}
\bibliography{refs}
\appendix
\section{Related Work}
\subsection{RLVR}
Reinforcement learning with verifiable feedback(RLVR) has emerged as a
commonly used paradigm for aligning LLMs with task-specific
goals such as problem-solving or code-generation.   
It \cite{shao2024deepseekmathpushinglimitsmathematical, wang2025reinforcementlearningreasoninglarge} can be seen as a simplified form of bootstrapping LM reasoning \cite{zelikman2024quietstarlanguagemodelsteach} or a simpler form of RL with execution feedback \cite{gehring2024rlef}, in which one uses answer matching or constraint verification as a binary signal to train large language models. RLVR allows an LLM to learn to reason better, especially in verifiable domains such as mathematics and programming. 

\subsection{Scaling Test-Time Compute}
Test-time compute scaling improves the performance of language models by allocating additional inference budget combining strategies such as search and voting. These approaches increase the diversity of candidate answers and, when combined with majority voting or verifier models, allow systems to reason for longer and arrive at more accurate solutions. This paradigm enables smaller open models to match or even surpass larger ones, as demonstrated in OpenAI’s o1 \cite{openai2024openaio1card}. Test-time scaling also integrates naturally with RLVR, since stronger verifiers directly improve inference-time selection \cite{snell2024scalingllmtesttimecompute, sareen2025putting,lambert2025tulu3pushingfrontiers}.

Verifiers \cite{guan2024searchverifyfeedbackgeneration} play a central role in these scaling-test time strategies, guiding inference and post-training optimization with scalable, generalizable supervision signals. However, as sample sizes grow, the effectiveness of verifier-guided approaches can decline \cite{yu2025scalingflawsverifierguidedsearch} due to misranking and pruning errors, particularly on challenging or out-of-distribution tasks, highlighting the need for improved calibration and hybrid candidate selection for robust scaling \cite{li2025verifybenchsystematicbenchmarkevaluating,yu2025scalingflawsverifierguidedsearch}.

\subsection{Efficient Verifiers}
Several recent approaches explicitly target efficiency of the verifier used in RLVR, \citet{chen2025xverifyefficientanswerverifier} introduces a small verifier (0.5B–3B params) that rivals closed source models like GPT-4o in answer equivalence detection across 10+ LLMs and datasets. Building on this, \citet{xu2025tinyvreducingfalsenegatives} reduces false negatives in rule-based verification by training a compact LLM verifier on  false negatives and positives, enabling more accurate reward estimates while maintaining computational efficiency when used in tandem with rule-based methods like \citet{cui2025processreinforcementimplicitrewards}. Other methods improve data efficiency by training on feedback-rich rationales  or by co-training verifiers with generators in reinforcement learning frameworks \citep{sareen2025putting, cui2025processreinforcementimplicitrewards}. Our verifier extends on this work by demonstrating that small yet robust RL verifiers can deliver high calibration and accuracy while reducing computational cost.

\section{Dataset Preparation}
\label{sec:datapreparation}

\textbf{LLM Response Generation}
We use the Curator package from Bespoke Labs for high-throughput generation of LLM responses to these questions, by sending asynchronous inference requests \cite{curator2025} to an online inference server from vLLM \cite{kwon2023efficientmemorymanagementlarge}. This combination of tools handles continuous batching and kernel optimizations to maximize GPU utilization. 

\textbf{Initial dataset selection}
Our dataset is prepared starting with the math and coding subsets from the Skywork-OR1-RL-Data dataset \cite{he2025skywork, skywork-or1-2025}, particularly the subsets sourced from NuminaMath \cite{numina_math_datasets}, TACO \cite{li2023taco} and LeetCode \cite{xia2025leetcodedatasettemporaldatasetrobust}.

\textbf{Existing Dataset and Software Issues}
Though these datasets are meant to be easily verifiable, we still find issues in the data when trying to check correctness. The Math-Verify package by default returns False when there is an internal error in the correctness check, rather than exposing the error. This is misleading, as a False correctness check should indicate that the model had a valid response that was different than a valid ground-truth. Instead we find that some of the errors are because Math-Verify cannot handle certain forms of equations found in the dataset. We adjusted the implementation to account for this and removed examples that created errors.    

NuminaMath is also a noisy dataset, as it is created through optical character recognition (OCR) on pdf files of heterogeneous formats at scale. We refer to the Skywork Open Reasoner Technical Report for full details of their initial filtering, but it generally involved heuristics to remove proof-based and other open-ended questions; removed questions with 0\% or 100\% success-rate from a moderate-sized model; and deduplicated questions based on embedding similarity. They also perform LLM-as-a-judge quality assessment of questions with LLama-3.3-70B-Instruct and Qwen2.5-72B-Instruct, with 16 samples per model, and retain only questions with >9 positive judgments on a binary scale. 

\textbf{Additional Filtering Contributions}
To further increase the quality of correctness signal provided to our verifier, we perform additional filtering. We eliminate questions that have multiple answers, as the comparison becomes more challenging. We remove questions that have answers in multiple choice format (such as A, B, C or D), as we find that the dataset is inconsistent where the question asks for the multiple choice letter, but contains as ground-truth the \textit{option} text from that choice, and vice-versa. We finally use sympy \cite{Meurer_SymPy_symbolic_computing_2017} to parse the ground-truth answer and determine if the answer evaluates to a single number. This checks that there are no free symbols or undefined functions; floats and similar are retained.   

The coding datasets are cleaner, and we simply use the Skywork version which checked that all test cases pass in the original solutions, and remove samples with empty, incomplete, or corrupted test cases.  

\textbf{Final Verification Methods}
For final math correctness checking, we use the Math-Verify package \cite{Kydlicek_Math-Verify_Math_Verification}, where the ground-truth answer is formatted into a latex environment before parsing, and extract response from within \textbackslash boxed\{\} delimiters for comparison (we request the use of \textbackslash boxed\{\} in the prompt). For code, we execute the test cases from the dataset on the generated code in a sandbox environment \cite{curator2025}. All test cases must pass for a code sample to be deemed correct.

\section{Calibrated Ratings}
\label{sec:verifieranalysis}

\begin{figure}[htb]
\centering
\includegraphics[width=\textwidth,keepaspectratio]{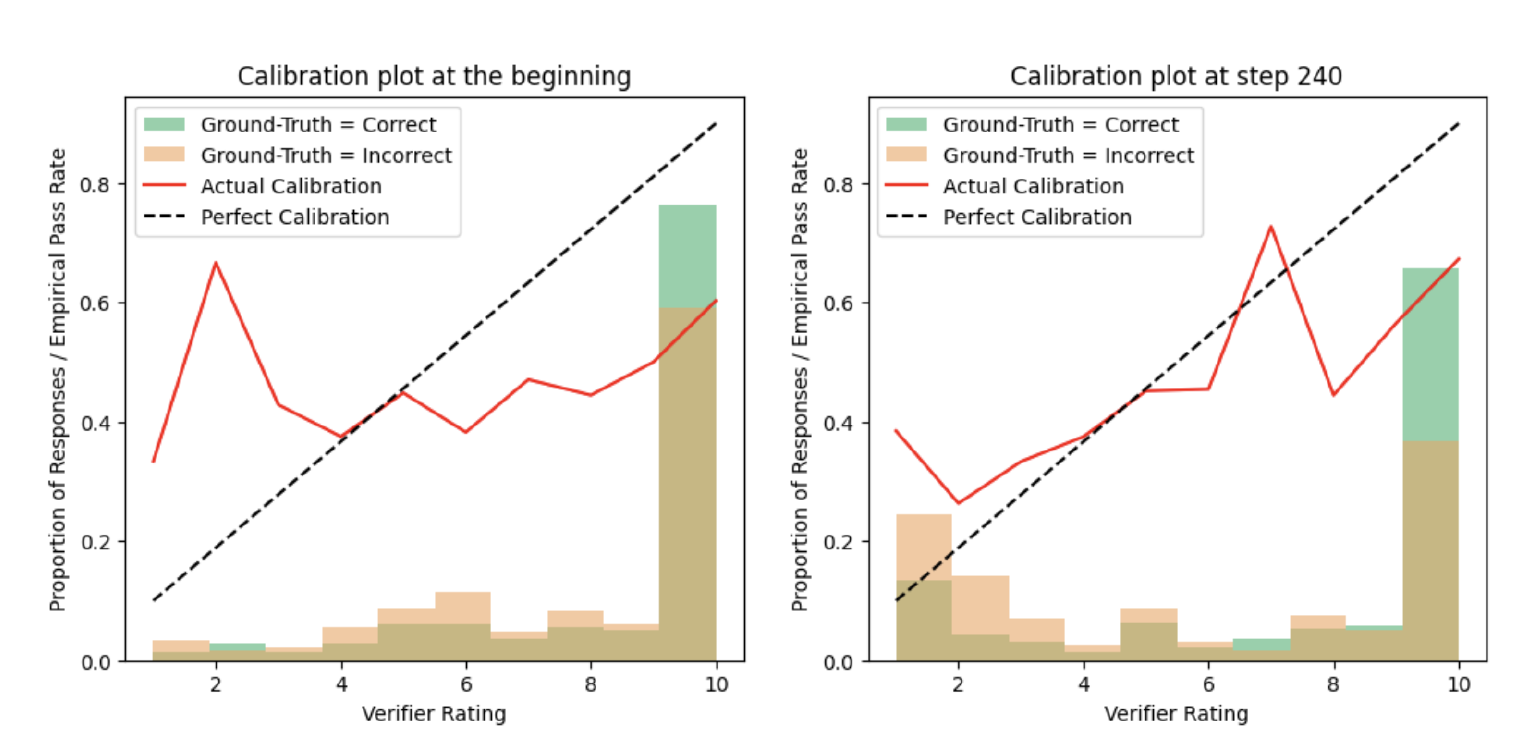}
\caption{\textbf{Left}: baseline \textsc{Qwen3-8B}, where ratings are skewed toward 10 and show little correlation with correctness. \textbf{Right}: trained verifier, which produces a broader spread of ratings and improved calibration, with accuracy (red) aligning more closely with the ideal (black).}
\label{fig:calibration_plot} 
\end{figure}

As illustrated in \cref{fig:calibration_plot}, the ratings produced by the trained verifier exhibit strong alignment with calibrated confidence estimates. The left panel reports results from the baseline \textsc{Qwen3-8B} model, where ratings are overwhelmingly concentrated at the maximum value of 10, largely independent of response correctness. The reference black line denotes the expected proportion of correct responses for a perfectly calibrated verifier, whereas the red line reflects empirical accuracy within each rating bucket. The flat red line at approximately 50\% demonstrates that the baseline model offers little discriminatory power across rating levels. By contrast, the right panel shows the trained verifier, which distributes ratings more broadly across the scale and, crucially, yields accuracy curves (red line) that track much more closely with the calibration reference (black line). This indicates that training substantially improves the reliability and interpretability of verifier scores as confidence measures.

\vspace{1.7in}
\section{Repeated Incorrect answers}
\label{sec:biasedanswers}

\begin{figure}[htb]
\centering
\includegraphics[width=0.9\textwidth,keepaspectratio]{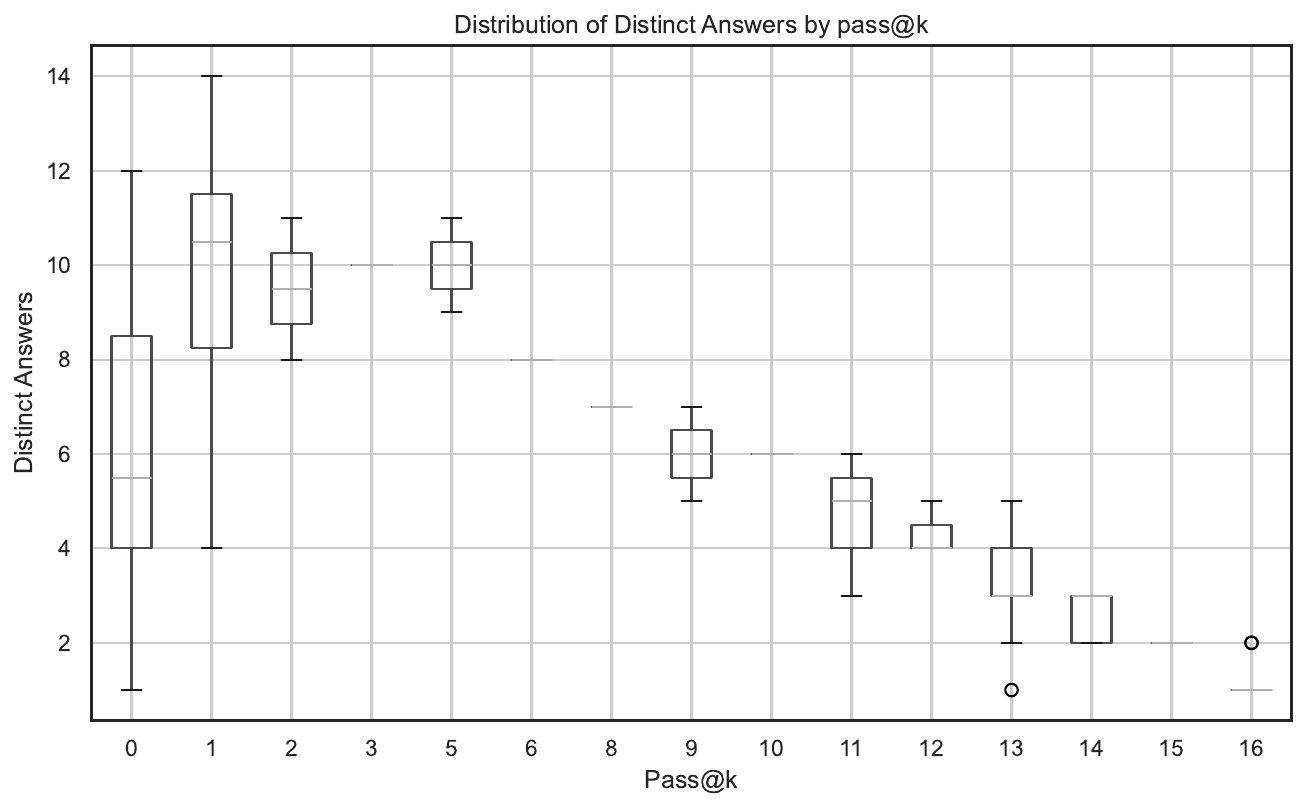}
\caption{\textbf{Model Bias Analysis} This figure illustrates the response diversity of \textsc{Qwen3-8B} over 16 generation attempts on problems of varying difficulty. For moderately complex problems (pass@k > 5), the model produces mostly correct solutions or diverse incorrect solutions (Y-axis count). In stark contrast, when faced with a difficult problem that yields no correct answers (pass@16 == 0), the model repeatedly generates the same incorrect solution, demonstrating a collapse into a narrow, biased failure mode.}
\label{fig:distinct_answers} 
\end{figure}

\end{document}